# Planning with Partially Observable Markov Decision Processes: Advances in Exact Solution Method

Nevin L. Zhang and Stephen S. Lee
Department of Computer Science, Hong Kong University of Science & Technology
{lzhang, sslee}@cs.ust.hk

## Abstract

There is much interest in using partially observable Markov decision processes (POMDPs) as a formal model for planning in stochastic domains. This paper is concerned with finding optimal policies for POMDPs. We propose several improvements to incremental pruning, presently the most efficient exact algorithm for solving POMDPs.

## 1 Introduction

*Partially observable Markov decision processes* (POMDPs) model sequential decision making problems where effects of actions are nondeterministic and the state of the world is not known with certainty. A POMDP consists of (1) a set $\mathcal{S}$ of possible states of the world, which is assumed to be finite in this paper, (2) a finite set $\mathcal{A}$ of possible actions, (3) a finite set $\mathcal{O}$ of possible observations. At each point in time, the world is in one of the possible states. An agent receives an observation $o$ according to an *observation probability* $P(o|s, a_-)$, which depends on the current state $s$ of the world and the action $a_-$ just executed. The minus sign in the subscript indicates the previous time point. The agent also chooses and executes an action. After an action $a$ is executed, the agent receives an immediate reward $r(s, a)$ and the world probabilistically moves into another state $s_+$ according to a *transition probability* $P(s_+|s, a)$. The plus sign in the subscript indicates the next time point.

The agent chooses actions based on its knowledge about the state of the world, which is summarized by a probability distribution over $\mathcal{S}$. The probability distribution is sometimes called a *belief state*. Let $b$ be the current belief state. If the agent observes $o_+$ after taking action $a$, then its next belief state $b_+$ is given by $b_+(s_+) = c\sum_s P(s_+, o_+|s, a)b(s)$, where $P(s_+, o_+|s, a) = P(o_+|s_+, a)P(s_+|s, a)$ and $c$ is the renormalization constant.

A *policy* maps each belief state to an action. A policy is *optimal* if it maximizes the expected long-term discounted reward. Value iteration is a standard way for finding policies that are arbitrarily close to optimal. It begins with an arbitrary initial function $V_0^*(b)$ of belief states and iterates using the following equation

$$V_t^*(b) = max_a[r(b,a) + \lambda \sum_{o_+} P(o_+|b,a)V_{t-1}^*(b_+)],$$

where $P(o_+|b,a) = \sum_{s,s_+} P(o_+, s_+|s,a)b(s)$ is the probability of observing $o_+$ after executing action $a$ in belief state $b$ and where $\lambda$ ($0<\lambda<1$) is a discount factor. Value iteration terminates when the *Bellman residual* $max_b|V_t^*(b) - V_{t-1}^*(b)|$, where the maximum is taken over all possible belief states, falls below a predetermined threshold $\epsilon$. An policy is then obtained through one step lookahead (e.g. Cassandra 1994).

Since there are uncountably infinite many belief states, value iteration cannot be carried out explicitly. Fortunately, it can be carried out implicitly. Sondik (1971) has shown that if there exists a finite set $\mathcal{V}_t$ of functions of $s$, henceforth called *vectors*, that *represents* $V_t^*$ in the sense that for all belief states $b$

$$V_t^*(b) = max_{\alpha \in \mathcal{V}_t} \sum_s \alpha(s)b(s),$$

then there exists a finite of vectors that represents $V_{t+1}^*$. If one begins with a function $V_0^*$, say 0, that can be represented by a finite set of vectors, then every $V_t^*$ can be represented by a finite set of vectors. Instead of computing $V_t^*$ explicitly, one can compute a set $\mathcal{V}_t$ of vectors that represents $V_t^*$.

The process of obtaining a minimal set of vectors that represents $V_{t+1}^*$ from a minimal set of vectors that represents $V_t^*$ is called *dynamic-programming update* (Littman et al 1995). It is of fundamental importance to POMDPs. Previous algorithms for dynamic-programming updates include one-pass (Sondik 1971), exhaustive (Monahan 1982), Lark (White 1991), linear support (Cheng 1988), witness (Littman *et al* 1995), and incremental pruning (Zhang and Liu 1997). Incremental pruning is the simplest among all those algorithms and preliminary experiments have shown that it is also the most efficient (Cassandra *et al* 1997).

This paper proposes a number of improvements to incremental pruning. We will begin with a formal statement of the dynamic-programming update problem (Section 2) and a brief review of incremental pruning (Section 3). We will then introduce the improvements (Sections 4 and 5) and discuss the pros and cons. Experimental results will be presented in Section 6 and conclusions provided in Section 7.

## 2   Dynamic-programming updates

To formally state the dynamic-programming update problem, we need several operations on and concepts about sets of vectors. Suppose $\mathcal{W}$ and $\mathcal{X}$ are two sets of vectors over the state space. The *cross sum* of $\mathcal{W}$ and $\mathcal{X}$ is a new set of vectors given by

$$\mathcal{W} \bigoplus \mathcal{X} = \{\alpha + \beta | \alpha \in \mathcal{W}, \beta \in \mathcal{X}\}.$$

It is evident that the cross sum operation is commutative and associative. Hence one can talk about the cross sum of more than two sets of vectors. Let $f(s_+, s)$ be a function of $s_+$ and $s$. The *matrix multiplication* of $\mathcal{W}$ and $f$ is a new set of vectors given by

$$\mathcal{W} * f = \{\beta | \exists \alpha \in \mathcal{W} \text{ s.t. } \beta(s_+) = \sum_{s_+} \alpha(s_+) f(s_+, s) \forall s_+\}$$

For any number $\lambda$, define $\lambda \mathcal{W} = \{\lambda \alpha | \alpha \in \mathcal{W}\}$.

A subset $\mathcal{W}'$ of $\mathcal{W}$ is a *covering* of $\mathcal{W}$ if for any belief state $b$, there exists $\alpha' \in \mathcal{W}'$ such that $\alpha'.b \geq \alpha.b$ for all $\alpha \in \mathcal{W}$. Here $\alpha'.b$ and $\alpha.b$ are the inner products of $\alpha'$ and $\alpha$ with $b$. A covering of $\mathcal{W}$ is *parsimonious* if none of its proper subsets are coverings of $\mathcal{W}$. If $\mathcal{W}$ is a parsimonious covering of itself, we say that $\mathcal{W}$ is *parsimonious*.

Let $\mathcal{B}$ be the set of all belief states. The *witness region* $R(\alpha, \mathcal{W})$ and the *closed witness region* $\overline{R}(\alpha, \mathcal{W})$ of a vector $\alpha \in \mathcal{W}$ w.r.t $\mathcal{W}$ are subsets of $\mathcal{B}$ respectively given by

$$R(\alpha, \mathcal{W}) = \{b \in \mathcal{B} | \alpha.b > \alpha'.b \quad \forall \alpha' \in \mathcal{W} \setminus \{\alpha\}\},$$

$$\overline{R}(\alpha, \mathcal{W}) = \{b \in \mathcal{B} | \alpha.b \geq \alpha'.b \quad \forall \alpha' \in \mathcal{W} \setminus \{\alpha\}\}.$$

It can be proved (Littman *et al* 1995) that $\mathcal{W}$ has a unique parsimonious covering and it is given by

$$PC(\mathcal{W}) = \{\alpha | R(\alpha, \mathcal{W}) \neq \emptyset\}.$$

A point in $R(\alpha, \mathcal{W})$ is called a *witness point* for $\alpha$ because it testifies to the fact that $\alpha$ is in the parsimonious covering $PC(\mathcal{W})$. A point in $\overline{R}(\alpha, \mathcal{W}) \setminus R(\alpha, \mathcal{W})$ is called a *boundary point* for $\alpha$.

Going back to value iteration, suppose $\mathcal{V}$ is a set of vectors that represents the function $V_t^*$. Given action $a$ and observation $o_+$, $P(o_+, s_+|s, a)$ is a function of $s_+$ and $s$. Moreover $r(s, a)$ is a function of $s$ and hence can be viewed as a vector over the state space. Define

$$\mathcal{V}_{a,o_+} = \lambda \mathcal{V} * P(o_+, s_+|s, a),$$
$$\mathcal{V}_a = \bigoplus_{o_+} \mathcal{V}_{a,o_+}, \quad \mathcal{U} = \cup_a [\{r(s,a)\} \bigoplus \mathcal{V}_a]. \quad (1)$$

Sondik (1971) has shown that the set $\mathcal{U}$ of vectors represents the function $V_{t+1}^*$. Consequently, the dynamic-programming update problem can be formally stated as follows:

> Given a parsimonious set $\mathcal{V}$ of vectors, find the parsimonious covering for the set $\mathcal{U}$ given by equation (1).

## 3   Incremental pruning

This section reviews the incremental pruning algorithm. We begin by considering the parsimonious covering of the cross sum of two sets $\mathcal{W}$ and $\mathcal{X}$ of vectors. For any $\alpha \in \mathcal{W}$ and any $\beta \in \mathcal{X}$, the vector $\alpha + \beta$ appears in the parsimonious covering $PC(\mathcal{W} \bigoplus \mathcal{X})$ if and only if the witness region $\alpha$ intersects with that of $\beta$, i.e. when $R(\alpha, \mathcal{W}) \cap R(\beta, \mathcal{X}) \neq \emptyset$. Consequently, the parsimonious covering $PC(\mathcal{W} \bigoplus \mathcal{X})$ can be obtained using the following procedure. The name CSP stands for cross-sum-pruning.

> Procedure CSP($\mathcal{W}, \mathcal{X}$):
> 1. $\mathcal{Y} \leftarrow \emptyset$.
> 2. For each $\alpha \in \mathcal{W}$ and each $\beta \in \mathcal{W}$,
> 3.     If $R(\alpha, \mathcal{W}) \cap R(\beta, \mathcal{X}) \neq \emptyset$, $\mathcal{Y} \leftarrow \{\alpha + \beta\} \cup \mathcal{Y}$.
> 4. Return $\mathcal{Y}$.

Whether the two witness regions $R(\alpha, \mathcal{W})$ and $R(\beta, \mathcal{X})$ intersect can be determined by solving the following linear program.

> Maximize: $x$.
> Constraints:
>   $\alpha.b \geq x + \alpha'.b$ for all other $\alpha' \in \mathcal{W}$
>   $\beta.b \geq x + \beta'.b$ for all other $\beta' \in \mathcal{X}$
>   $\sum_s b(s) = 1$, $b(s) \geq 0$ for all $s \in \mathcal{S}$

This linear program is always feasible; the first two groups of constraints are all satisfied for any belief state $b$ when $x$ is small enough. The witness region of $\alpha$ intersects with that of $\beta$ if and only if the solution for $x$ is positive.

Next consider the parsimonious covering of the cross sum of a list of sets of vectors $\mathcal{W}_1, \mathcal{W}_2, \ldots, \mathcal{W}_m$. For any $k$ such that $1 \leq k \leq m$, $PC(\bigoplus_{i=1}^{k} \mathcal{W}_i) \oplus \mathcal{W}_{k+1}$ is a covering of $\bigoplus_{i=1}^{k+1} \mathcal{W}_i$ and hence its parsimonious covering is the same as that of the latter. This leads to the following procedure for computing $PC(\bigoplus_{i=1}^{m} \mathcal{W}_i)$:

> Procedure IP($\{\mathcal{W}_i : i = 1, \ldots, m\}$):
> 1. $\mathcal{Y} \leftarrow \mathcal{W}_1$.
> 2. **For** $i = 2$ to $m$, $\mathcal{Y} \leftarrow \text{CSP}(\mathcal{Y}, \mathcal{W}_i)$.
> 3. Return $\mathcal{Y}$.

This procedure is named *incremental pruning* (IP) because pruning takes place while performing cross sums rather than after all the cross sums.

Suppose there are $m$ possible observations and enumerate them as 1, 2, ..., $m$. Applying IP to $\{\mathcal{V}_{a,o_+}|o_+=1,2,\ldots,m\}$), one obtains the parsimonious covering $PC(\mathcal{V}_a)$ of the set $\mathcal{V}_a$ defined in (1). The union $\cup_a[\{r(s,a)\} \oplus PC(\mathcal{V}_a)]$ is a covering of $\mathcal{U}$ and hence its parsimonious covering is same as that of the latter. This leads to the following algorithm for dynamic-programming updates:

> Procedure DP-Update($\mathcal{V}$):
> 1. For each $a$, $\mathcal{W}_a \leftarrow \text{IP}(\{\mathcal{V}_{a,o_+}|o_+ = 1, 2, \ldots, m\})$.
> 2. Return $PC(\cup_a[\{r(s,a)\} \oplus \mathcal{W}_a])$.

We also use the term incremental pruning to refer to the above algorithm for dynamic-programming updates. It does not specify which method one should use at line 2 to find the parsimonious covering of $\cup_a[\{r(s,a)\} \oplus PC(\mathcal{V}_a)]$. A popular choice is Lark's algorithm[1] (White 1991).

## 4 Improvements to incremental pruning

In DP-Update, IP is called once for each possible action to find the parsimonious covering of the cross sum of $m$ sets of vectors, where $m$ is the number of possible observations. In the process, CSP is called $m-1$ times. When calculating the parsimonious covering of two sets $\mathcal{W}$ and $\mathcal{X}$ of vectors, CSP solves $|\mathcal{W}||\mathcal{X}|$ linear programs and each linear program has $|\mathcal{W}|+|\mathcal{X}|+n-1$ constraints, where $n$ is the number of possible states.

---

[1] One can also apply Lark's algorithm directly to $\mathcal{U}$. This is, however, very inefficient since the size of $\mathcal{U}$ is exponential in the number of observations.

The restricted region variation of incremental pruning (Cassandra *et al* 1997) reduces the numbers of constraints in some of the linear programs by incorporating the idea behind Lark's algorithm into CSP. A number of linear programs can also be saved by exploiting the following fact. Suppose we know a witness point $b$ for a vector $\alpha$ in $\mathcal{W}$. If $b$ also happens to be a witness point for a vector $\beta \in \mathcal{X}$, then the witness regions of $\alpha$ and $\beta$ must intersect. This conclusion is reached without solving a linear program.

The section introduces four new improvements. The first reduces the number of calls to IP; the second and the third improvements reduces the number of linear programs and the numbers of constraints in the linear programs respectively; and the fourth improvement reformulates the linear programs so that they yield more information and uses the information to reduce the number of linear programs.

### 4.1 Reducing the number of calls to IP

Actions can be classified into those that gather information and those that achieve goals. In robot path planning, `move-forward`, `turn-left`, and `turn-right` are goal-achieving actions while `looking-around` is an information-gathering action. It is often the case that two different goal-achieving actions $a_1$ and $a_2$ have identical observation probabilities, i.e. $P(o_+|s_+,a_1)=P(o_+|s_+,a_2)$. This fact can be exploited to reduce the number of calls to IP.

For any action $a$ and any observation $o_+$, define

$$\mathcal{V}'_{a,o_+} = \{\beta | \exists \alpha \in \mathcal{V} \text{ s.t. } \beta(s_+)=\lambda\alpha(s_+)P(o_+|s_+,a)\forall s_+\},$$

$$\mathcal{V}'_a = \bigoplus_{o_+} \mathcal{V}'_{a,o_+}. \qquad (2)$$

Comparing these definitions with the ones given in (1), one can easily see that $\mathcal{V}_a = \mathcal{V}'_a * P(s_+|s,a)$. Moreover, $PC(\mathcal{V}'_a) * P(s_+|s,a)$ is a covering of $\mathcal{V}_a$. We can hence modify DP-Update as follows: For each action $a$, obtain the parsimonious covering of $\mathcal{V}'_a$ by applying IP to $\{\mathcal{V}'_{a,o_+}|o_+=1,2,\ldots,m\}$). Then apply Lark's algorithm to the union $\cup_a[\{r(s,a)\} \oplus [PC(\mathcal{V}'_a) * P(s_+|s,a)]]$. The result is still the parsimonious covering of $\mathcal{U}$.

Given $a$, $P(s_+|s,a)$ can be viewed as an $n \times n$ matrix. If the matrix is invertible, then $PC(\mathcal{V}'_a) * P(o_+|s_+,a)$ is also the parsimonious covering of $\mathcal{V}_a$. If the matrix is not invertible, $PC(\mathcal{V}'_a) * P(o_+|s_+,a)$ might be non-parsimonious. When this is the case, the above modification leaves more work to Lark's algorithm. However, there is a big advantage. If $P(o_+|s_+,a_1)=P(o_+|s_+,a_2)$, then $\mathcal{V}'_{a_1}=\mathcal{V}'_{a_2}$. Consequently, the computations for obtaining the parsimonious coverings $PC(\mathcal{V}_{a_1})$ and $PC(\mathcal{V}_{a_2})$ can be shared. The number of calls to IP is thereby reduced.

### 4.2 Reducing the number of linear programs in CSP

When computing the parsimonious covering $PC(\mathcal{W} \oplus \mathcal{X})$, CSP solves a linear program for each pair $(\alpha, \beta)$ of vectors $\alpha \in \mathcal{W}$ and $\beta \in \mathcal{X}$ to determine whether their witness regions intersect. This subsection explains how some of the linear programs can be saved if we know a witness or boundary point for each vector in $\mathcal{W}$ and the neighboring relationships among witness regions of vectors in both $\mathcal{W}$ and $\mathcal{X}$. We will show later how such knowledge can be made available through proper book keeping and some additional computations.

Because of the constraint $\sum_s b(s)=1$, closed witness regions are polytops in the $n-1$ dimensional space. For any two vectors $\alpha$ and $\alpha'$ in $\mathcal{W}$, the intersection of the closed witness regions $\overline{R}_{(\alpha,\mathcal{W})}$ and $\overline{R}_{(\alpha',\mathcal{W})}$ is a polytop of dimension less than $n-1$. If the polytop is of dimension $n-2$, we say that the closed witness regions are *neighbors*. When this is the case, we also say that the vectors $\alpha$ and $\alpha'$ are *neighbors* in $\mathcal{W}$. It is easy to see that the set of all neighbors of $\alpha$ is the minimum subset $\mathcal{W}'$ of $\mathcal{W}$ such that $R(\alpha, \mathcal{W}) = R(\alpha, \mathcal{W}')$.

If we know a witness or boundary point for each vector in $\mathcal{W}$ and the neighboring relationships among vectors in $\mathcal{W}$ and among vectors in $\mathcal{X}$, then CSP can be modified as follows: To initialize, set $\mathcal{Y}=\emptyset$. For each $\alpha \in \mathcal{W}$,

1. Find all vectors $\beta$ in $\mathcal{X}$ that has maximum inner product with the witness or boundary point of $\alpha$.

2. For each such vector $\beta$, determine whether the witness regions $R(\alpha, \mathcal{W})$ and $R(\beta, \mathcal{X})$ intersect.

3. If they do, add $\alpha+\beta$ to $\mathcal{Y}$ and repeat 2 for all the neighbors of $\beta$ that have not been considered.

It can be proved that the set $\mathcal{Y}$ is $PC(\mathcal{W} \oplus \mathcal{X})$ when the procedure terminates.

For a given vector $\alpha \in \mathcal{W}$, a vector $\beta \in \mathcal{X}$ is examined by the above procedure if and only if its witness region or that of one of its neighbors intersect with the witness region of $\alpha$. The number of such $\beta$'s is much smaller than the total number of vectors in $\mathcal{X}$ when the sets $\mathcal{W}$ and $\mathcal{X}$ are large (which is usually the case) because then the witness regions are small.

### 4.3 Reducing the number of constraints

Consider the linear program in Section 3. Replace the first two groups of constraints by the following:

$\alpha.b \geq x+\alpha'.b$ for all neighbors $\alpha'$ of $\alpha$,
$\beta.b \geq x+\beta'.b$ for all neighbors $\beta'$ of $\beta$.

The solution for $x$ is positive in the modified linear program if and only if this is the case in the original linear program. However, the modified linear program is easier to solve because it contains fewer constraints.

### 4.4 Reformulating linear programs

The linear program in Section 3 enables us to determine whether the two witness regions $R(\alpha, \mathcal{W})$ and $R(\beta, \mathcal{X})$ intersect. When they do, the linear program also gives us a witness point for the vector $\alpha+\beta \in PC(\mathcal{W} \oplus \mathcal{X})$, which is the belief point $b$ that allows $x$ to take its maximum value. We will refer to this point as the *maximum point* of the linear program. This subsection reformulates the linear program so that it gives us more information and hopefully helps us saving some linear programs.

Here is the reformulated linear program:

Maximize: $x$.
Constraints:
$\quad \alpha.b \geq x+\alpha'.b$ for all neighbors $\alpha'$ of $\alpha$
$\quad \beta.b \geq \beta'.b$ for all neighbors $\beta'$ of $\beta$
$\quad \sum_s b(s) = 1$, $b(s) \geq 0$ for all $s \in \mathcal{S}$

In addition to the fact that it incorporates the improvement described in the previous subsection, the reformulated linear program differs from the original one only in the absence of $x$ from the second group of constraints.

The reformulated linear program is infeasible if and only if the closed witness region of $\beta$ is empty. In this case, $\beta$ should be pruned from $\mathcal{X}$. From now on, we assume the linear program is feasible.

The witness regions of $\alpha$ and $\beta$ intersect if and only if the solution for $x$ in the reformulated linear program is positive. When this is the case, the maximum point $b$ is either a witness point or a boundary point for the vector $\alpha+\beta \in PC(\mathcal{W} \oplus \mathcal{X})$. It is a boundary point of $\alpha+\beta$ if and only if there exist neighbors $\gamma$ of $\beta$ such that $\beta.b=\gamma.b$. When this is the case, the witness region of any such neighbor $\gamma$ of $\beta$, if not empty, must intersect with that of $\alpha$. Thus, without solving additional linear programs, we know that the vector $\alpha+\gamma$ is in $PC(\mathcal{W} \oplus \mathcal{X})$ and $b$ is one of its boundary points.

Now consider the case when the solution for $x$ is not positive. Here the witness regions of $\alpha$ and $\beta$ do not intersect. Define region $\overline{R}_x(\alpha, \mathcal{W}) = \{b \in \mathcal{B} | \alpha.b \geq x+\alpha'.b$ for all neighbors $\alpha'$ of $\alpha\}$. It grows as $x$ decreases. Thus $x$ achieves its maximum possible value when the region first touches the closed region $\overline{R}(\beta, \mathcal{X})$. Consequently, the maximum point $b$ of the reformulated linear program must be a boundary point of $\beta$ and there must exist neighbors $\gamma$ of $\beta$ such that $\beta.b=\gamma.b$. Suppose $b$ is a witness point for some $\alpha'\in\mathcal{W}$. If the witness region of $\beta$ is not empty, then it intersects with that of $\alpha'$. Hence the vector $\alpha'+\beta$ is in $PC(\mathcal{W}\oplus\mathcal{X})$ and $b$ is one of its boundary point. Similarly, for any neighbor $\gamma$ of $\beta$ such that $\gamma.b = \beta.b$ and $R(\gamma,\mathcal{X})\neq\emptyset$, the vector $\alpha'+\gamma$ is in $PC(\mathcal{W}\oplus\mathcal{X})$ and $b$ is one of its boundary points. Again, we know all those without solving additional linear programs.

## 5 Facilitating the improvements

The improvement described in Subsection 4.1 reduces the number of calls to IP. We shall refer to it as an *IP-reduction technique*. The improvements introduced in Subsections 4.2-4.4, on the other hand, reduce the number of linear programs that CSP has to solve and the numbers of constraints in those linear programs. We shall refer to them as *LP-reduction techniques*.

The IP-reduction technique calls for the computation of $PC(\mathcal{V}'_a)$ by applying IP to $\{\mathcal{V}'_{a,o_+}|o_+=1,2,\ldots,m\})$. In the process, CSP is called $m-1$ times. In the $k$th call ($1\leq k\leq m-1$), the inputs to CSP are $PC(\bigoplus_{i=1}^{k}\mathcal{V}'_{a,o_+=i})$ and $\mathcal{V}'_{a,o_+=k+1}$. To facilitate the LP-reduction techniques, we need a witness or boundary point for each vector in the first set and the neighboring relationships among vectors in both sets. This section show how those can be made available through proper book keeping and some additional computations.

### 5.1 Inheritance of neighboring relationships

Assume we know the exact neighboring relationships among vectors in the set $\mathcal{V}$. This subsection shows how one can, through proper book keeping, identify most pairs of vectors in $PC(\bigoplus_{i=1}^{k}\mathcal{V}'_{a,o_+=i})$ and $\mathcal{V}'_{a,o_+=k+1}$ that are not neighbors. A method for finding the exact neighboring relationships among vectors in $\mathcal{V}$ will be described later.

Pairs of vectors that are not identified as non-neighbors will simply regarded as neighbors. Treating non-neighbors as neighbors increases the complexities of the LP-reduction techniques. Fortunately, those techniques yield the correct results as long as neighbors are not mistaken as non-neighbors.

Consider two vectors $\beta$ and $\beta'$ in $\mathcal{V}'_{a,o_+}$. By the definition of $\mathcal{V}'_{a,o_+}$, there must exist vectors $\alpha$ and $\alpha'$ in $\mathcal{V}$ such that $\beta s_+=\alpha(s_+)P(o_+|s_+,a)$ and $\beta'(s_+)=\alpha'(s_+)P(o_+|s_+,a)$. Using the property of neighbors mentioned right after the concept was defined in Subsection 4.2, one can show that if $\alpha$ and $\alpha'$ are not neighbors in $\mathcal{V}$, then $\beta$ and $\beta'$ cannot be neighbors in $\mathcal{V}'_{a,o_+}$.

Next consider the parsimonious covering $PC(\mathcal{W}\oplus\mathcal{X})$ of two sets $\mathcal{W}$ and $\mathcal{X}$ of vectors. Each member of the covering can be written as $\alpha+\beta$, where $\alpha\in\mathcal{W}$ and $\beta\in\mathcal{X}$. The witness region of $\alpha+\beta$ w.r.t $PC(\mathcal{W}\oplus\mathcal{X})$ is simply the intersection $R(\alpha,\mathcal{W})\cap R(\beta,\mathcal{X})$ and hence is a subset of both $R(\alpha,\mathcal{W})$ and $(\beta,\mathcal{X})$. This fact implies that two vectors $\alpha+\beta$ and $\alpha'+\beta'$ in the covering cannot be neighbors if $\alpha$ and $\alpha'$ are not neighbors in $\mathcal{W}$ or $\beta$ and $\beta'$ are not neighbors in $\mathcal{W}$.

Now consider the case when $\alpha$ is a neighbor of $\alpha'$ and $\beta$ is a neighbor of $\beta'$. Suppose $\alpha+\beta$ and $\alpha'+\beta'$ are neighbors in $PC(\mathcal{W}\oplus\mathcal{X})$. Then the intersection of their closed witness regions is an $n-2$ dimensional polytop. Denote this polytop by $A$. Since the closed witness region of $\alpha+\beta$ is a subset of both $\overline{R}(\alpha,\mathcal{W})$ and $\overline{R}(\beta,\mathcal{X})$ and that of $\alpha'+\beta'$ is a subset of both $\overline{R}(\alpha',\mathcal{W})$ and $\overline{R}(\beta',\mathcal{X})$, the polytops $\overline{R}(\alpha,\mathcal{W})\cap\overline{R}(\alpha',\mathcal{W})$ and $\overline{R}(\beta,\mathcal{X})\cap\overline{R}(\beta',\mathcal{X})$ must lie in the same $n-2$ dimensional space as the polytop $A$. Since the polytop $\overline{R}(\alpha,\mathcal{W})\cap\overline{R}(\alpha',\mathcal{W})$ lies in the space $\{b\in\mathcal{B}|\alpha.b=\alpha'.b\}$ while the polytop $\overline{R}(\beta,\mathcal{X})\cap\overline{R}(\beta',\mathcal{X})$ lies in the space $\{b\in\mathcal{X}|\beta.b=\beta'.b\}$, it must be the case that $\alpha-\alpha'=c(\beta-\beta')$, where $c$ is some constant. When this is not the case, $\alpha+\beta$ and $\alpha'+\beta'$ cannot be neighbors.

### 5.2 Witness and boundary points

This subsection discusses how the need for a witness or boundary point for each vector in the set $PC(\bigoplus_{i=1}^{k}\mathcal{V}'_{a,o_+=i})$ can be facilitated. If $k>1$, the set is obtained by CSP from $PC(\bigoplus_{i=1}^{k-1}\mathcal{V}'_{a,o_+=i})$ and $\mathcal{V}'_{a,o_+=k}$. Consider the linear programs that CSP solves. If they are of the form given in Section 3, they produce witness points for vectors in $PC(\bigoplus_{i=1}^{k}\mathcal{V}'_{a,o_+=i})$ as by-products. If they are of the form given in Subsection 4.4, they yield witness or boundary points for each vector in $PC(\bigoplus_{i=1}^{k}\mathcal{V}'_{a,o_+=i})$ as by-products.

The rest of this subsection deals with the case when $k=1$. Here we need to find a witness or boundary point, if exists, for each vector in $\mathcal{V}'_{a,o_+=1}$. For notational simplicity, we consider the set $\mathcal{V}'_{a,o_+}$ for a general observation $o_+$.

Assume the set $\mathcal{V}$ is parsimonious and a witness or boundary point is known for each vector in $\mathcal{V}$. We can make this assumption because, to solve a POMDP, DP-Update needs to be called iteratively until a certain stopping criterion is met. At the first iteration, $\mathcal{V}$ typically consists of only one vector. Any belief state is a witness point for the vector. At later iterations, $\mathcal{V}$ is the output of the previous call to DP-Update. If DP-Update uses Lark's algorithm at line 2, witness or boundary points for vectors in $\mathcal{V}$ are computed as by-products.

It is obviously desirable to take advantage of the known witness or boundary points for vectors in $\mathcal{V}$ when finding witness or boundary points for vector in $\mathcal{V}_{a,o_+}$. Let $b$ be the known witness or boundary point for a vector $\alpha \in \mathcal{V}$. Define a new belief state $b'$ by setting $b'(s_+) = cb(s_+)/P(o_+|s_+, a)$ when $P(o_+|s_+, a) > 0$ and $b'(s_+) = 0$ otherwise, where $c$ is the renormalization constant. If $P(o_+|s_+, a) > 0$ for all possible values of $s_+$, then $b'$ must be a witness or boundary point for the vector $\beta \in \mathcal{V}'_{a,o_+}$ such that $\beta(s_+) = \alpha(s_+)P(o_+|s_+, a)$.

Now consider the case when $P(o_+|s_+, a) = 0$ for some possible values $s_+$. If there is a vector $\beta \in \mathcal{V}'_{a,o_+}$ such that $\beta.b' \geq \beta'.b'$ for all other $\beta' \in \mathcal{V}'_{a,o_+}$, then $b'$ is a witness or boundary point for $\beta$. This fact allows us to find witness or boundary points for some of the vectors. For vectors $\beta \in \mathcal{V}'_{a,o_+}$ whose witness or boundary points are not found this way, we solve the following linear program:

Maximize: $x$.
Constraints:
$\quad \beta.b \geq x + \beta'.b$ for all neighbors $\beta'$ of $\beta$ [2]
$\quad b(s_+) = 0$ for all $s_+$ such that $P(o_+|s_+, a) = 0$
$\quad \sum_s b(s_+) = 1$, $b(s_+) \geq 0$ for all $s_+ \in \mathcal{S}$

When the solution for $x$ is positive, the maximum point of the linear program is a witness point of $\beta$. When the solution for $x$ is not positive, the witness region of $\beta$ is empty. Hence $\beta$ can be pruned from $\mathcal{V}_{a,o_+}$.

### 5.3 Identifying neighboring relationships

This subsection shows how to find the neighboring relationships among vectors in $\mathcal{V}$. As mentioned earlier, $\mathcal{V}$ typically consists of only one vector the first time when DP-Update is called. The neighboring relationships are trivial in this case. At later calls, $\mathcal{V}$ is the output of the previous call to DP-Update. Consequently, it suffices to show how the neighboring relationships among vectors in the output $PC(\mathcal{U})$ of DP-Update can be found.

Use $\mathcal{W}_a$ to denote $\{r(s, a)\} \bigoplus [PC(\mathcal{V}'_a) * P(s_+|s, a)]$. Then $PC(\mathcal{U}) = PC(\cup_a \mathcal{W}_a)$. As discussed in Subsection 5.1, certain pairs of vectors in $PC(\mathcal{V}'_a)$ are known to be non-neighbors due to neighboring relationship inheritance. Those known relationships are in turn inherited by $\mathcal{W}_a$. Consider any two vectors $\beta$ and $\beta'$ in $\mathcal{W}_a$. They can be written as $\beta(s) = r(s, a) + \sum_{s_+} \alpha(s_+)P(s_+|s, a)$ and $\beta'(s) = r(s, a) + \sum_{s_+} \alpha'(s_+)P(s_+|s, a)$, where $\alpha$ and $\alpha'$ are vectors in $PC(\mathcal{V}'_a)$. It can be proved that $\beta$ and $\beta'$ cannot be neighbors in $\mathcal{W}_a$ if $\alpha$ and $\alpha'$ are not neighbors in $PC(\mathcal{V}'_a)$.

Consider a vector $\alpha \in PC(\mathcal{U})$. We find all its neighbors in two steps: first detect vectors that might be neighbors of $\alpha$ and then examine each of the potential neighbors to determine whether it is indeed a neighbor. There must exist action $a$ such that $\alpha \in \mathcal{W}_a$. The first step takes advantage of the known non-neighboring relationships among vectors in $\mathcal{W}_a$. It relies on the following fact. Suppose $\mathcal{N}_a$ is a list of vectors in $\mathcal{W}_a$ that potentially are neighbors of $\alpha$ in $\mathcal{W}_a$ and suppose $\mathcal{N}$ is a list of vectors in $PC(\mathcal{U})$ that potentially are neighbors of $\alpha$ in $PC(\mathcal{U})$. Then a vector $\beta$ in $PC(\mathcal{U}) \setminus \mathcal{N}$ can be a neighbor of $\alpha$ only if the following linear program has a positive solution for its objective function:

Maximize: $x$.
Constraints:
$\quad \alpha.b = \beta.b$
$\quad \alpha.b \geq x + \alpha'.b$ for all $\alpha' \in \mathcal{N}_a \cup \mathcal{N}$.
$\quad \sum_s b(s) = 1$, $b(s) \geq 0$ for all $s \in \mathcal{S}$

To find all potential neighbors of $\alpha$, initialize $\mathcal{N}$ to the empty set. Examine each vector $\beta$ in $PC(\mathcal{U}) \setminus \{\alpha\}$ using the above linear program. The linear program is skipped if $\alpha$ and $\beta$ are already known to be neighbors or non-neighbors. The vector $\beta$ is added to $\mathcal{N}$ once it is detected as a potential neighbor. After examing all the vectors, none of the vectors outside $\mathcal{N}$ can be neighbors of $\alpha$.

The second step first makes use of the belief points found when solving instances of the above linear program. Suppose $b$ is the belief point found when a vector $\beta$ is detected as a potential neighbor of $\alpha$. If $\beta.b > \beta'.b$ for any other potential neighbor $\beta'$, then $\beta$ must be a neighbor. If this is not the case, another linear program needs to be solved to determine whether $\beta$ is a neighbor of $\alpha$. This new linear program is the same as the one above except with the phrase "for all $\alpha' \in \mathcal{N}_a \cup \mathcal{N}$" replaced by "for all $\alpha' \in \mathcal{N}$". The vector $\beta$ is a neighbor of $\alpha$ if and only if the solution for $x$ in the new linear program is positive [3]. A potential neighbor is removed from the list $\mathcal{N}$ once it is found not to be a neighbor.

A couple of facts can be used to reduce the number of linear programs. First if two vectors in $PC(\mathcal{U})$ are from the same $\mathcal{W}_a$ and they are not neighbors in $\mathcal{W}_a$, then they cannot be neighbors in $PC(\mathcal{U})$ either. This is true because the witness region of a vector in $PC(\mathcal{U})$ must be a subset of its witness region in $\mathcal{W}_a$.

---

[2] (Non-)neighboring relationships among vectors in $\mathcal{V}'_{a,o_+}$ inherit from those among vectors in $\mathcal{V}$ in the way described in the previous section. Pairs of vectors that are not identified as non-neighbors are simply treated as neighbors.

[3] One might suggest to avoid the first step by regarding all other vectors as potential neighbors of $\alpha$. The problem with this alternative is that the resulting linear programs contain a large number of constraints.

Second if $PC(\mathcal{U})$ is obtained from $\cup_a \mathcal{W}_a$ using Lark's algorithm, then we have a witness or boundary point for each vector in $PC(\mathcal{U})$. If a belief point $b$ is a witness point for a vector $\alpha \in PC(\mathcal{U})$, then any vector $\beta \in PC(\mathcal{U})\setminus\{\alpha\}$ such that $\beta.b \geq \beta'.b$ for all other $\beta' \in PC(\mathcal{U})\setminus\{\alpha\}$ must be a neighbor of $\alpha$.

## 6 Discussions

The IP-reduction technique has no overhead while the LP-reduction techniques do. Our experiences indicate that the main overhead is the need to identify neighbors for vectors in $PC(\mathcal{U})$. In the neighbor detection method presented in Subsection 5.3, the second step does not take much time at all. The number of linear programs solved at the first step is upper bounded by $n(n-1)/2$, where $n$ is the number of vectors in $PC(\mathcal{U})$. The numbers of constraints in those linear programs are much smaller than $n$ when $n$ is large.

At the next time when DP-Update is called, new sets of vectors are constructed for various combinations of $o_+$ and $a$ by multiplying each vector in $PC(\mathcal{U})$ with $P(o_+|s_+,a)$. If $P(o_+|s_+,a)>0$ for all $s_+$, the corresponding new set of vectors is parsimonious. To compute the parsimonious covering of the cross sum of two such sets, the original CSP solves $n^2$ linear programs and each of them has $2(n-1)$ constraints. The overhead of the LP-reduction techniques at each call to DP-Update is hence upper bounded by the complexity of one call to CSP in the original incremental pruning algorithm. The LP-reduction techniques can significantly speed up dynamic-programming update because they drastically reduce the number of linear programs CSP has to solve and the numbers of constraints in those linear programs.

There are two cases where the incorporation of the LP-reduction techniques can be counter-productive. The first case is when the number of cross sums needed at each call to DP-Update is very small (e.g. no more than 3). As an extreme example, suppose there are only two possible observations and the observation probabilities for all actions are equal. Then one needs to perform only one cross sum at each call to DP-Update provided the IP-reduction technique is incorporated. Savings due to the LP-reduction techniques in this only cross sum cannot offset the costs of those techniques.

The second case is when the observations are very informative so that $P(o_+|s_+,a)>0$ only for a small number of possible states $s_+$. In this case, the sizes of the new sets mentioned above can be greatly reduced, be-fore the sets are fed to CSP, by pruning vectors that are pointwise dominated by others [4]. As a consequence, the cost incurred when identifying neighbors for vectors in $PC(\mathcal{U})$ might not be compensated by savings in the cross sums.

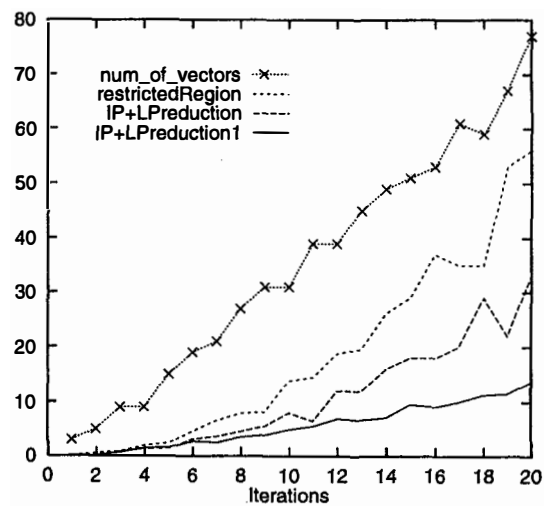

Figure 1: Comparisons between two variations of incremental pruning: restricted region and one that incorporates LP-reduction techniques.

The issue of exploiting informative observations is studied in detail in Zhang and Liu (1997). The LP-reduction techniques can be incorporated into the method.

## 7 Experiments

Preliminary experiments have been conducted to determine the effectiveness of the improvements. Due to time constraints, we have so far implemented only the two LP reduction techniques described in Subsections 4.2 and 4.3. Cassandra et al (1997) have shown that the restricted region variation of incremental pruning is significantly more efficient than plain incremental pruning. This section reports empirical comparisons between restricted region and a new variation of incremental pruning that incorporates the above two LP reduction techniques.

The tiger problem (Cassandra 1994) was used in the experiments. Figure 1 shows the times that the two algorithms took at the first twenty iterations. The number of vectors at the beginning of each iteration is also shown. We cut off at iteration twenty because thereafter the two algorithms, due to machine precision, produce different numbers of vectors.

---

[4]Pointwise pruning is computationally very cheap.

The curve IP+LPreduction depicts the total time, in CPU seconds, that the new variation of incremental pruning took, while IP+LPreduction1 depicts the total time minus the time spent in finding neighboring relationships. The difference between restrictedRegion and IP+LPreduction1 represent the gains of the LP reduction techniques, while that between IP+LPreduction and IP+LPreduction1 represent the overhead. We see that the gains are significant.

On the other hand, the overhead is also large. Fortunately, neighboring relationships need to be computed only once for each iteration. As a consequence, the overhead does not increase with the numbers of possible actions and observations, while the gains do. In the tiger problem, there are three possible actions and two possible observations and hence three cross sums are performed at each iteration. The net gains, i.e. the difference between restrictedRegion and IP+LPreduction, are not very significant in this case. If there were a large number of possible actions and observations, the net gains would be close to the difference between restrictedRegion and IP-LPreduction1.

It should be noted that the tiger problem has only two possible states. One implication is that the vectors that incremental pruning deals with are two dimensional and each vector can have at most two neighbors. Experiments are under way to determine the effectiveness of the two LP reductions techniques, as well as the other two improvements, on problems with larger state spaces.

## 8 Conclusions

Incremental pruning is presently the most efficient exact algorithm for finding optimal policies for POMDPs. It solves many linear programs. This paper proposes four improvements to incremental pruning. The first improvement reduces the number of linear programs by taking advantage of the fact that different actions sometimes have equal observation probabilities. The second and third improvements further reduce the number of linear programs and the numbers of constraints in the linear programs by exploiting neighboring relationships among witness regions. The fourth improvement reformulates the linear programs so that they provide us with more information and hence hopefully reduces the number of linear program even further. Preliminary experiments haves shown that the improvements can significantly speed up incremental pruning.

It is unlikely that exact algorithms by themselves can solve large POMDPs. A obvious future direction is to incorporate the ideas behind the exact methods into approximate algorithms. The techniques introduced in this paper can be easily incorporated into the approximate method proposed by Zhang and Liu (1997).

## Acknowledgements

The authors would like thank Anthony R. Cassandra, Robert Givan, and Michael Littman for valuable discussions and Weihong Zhang for comments on earlier versions of this paper.## References

Cassandra, A. R. (1994). Optimal polices for partially observable Markov decision processes. TR CS-94-14, Department of Computer Science, Brown University, Providence, Rhode Island 02912, USA.

Cassandra, A. R., Littman, M. L., & Zhang, N. L. (1997). Incremental pruning: A simple, fast, exact method for partially observable Markov decision processes. In *Proceedings of Thirteenth Conference on Uncertainty in Artificial Intelligence*, 54-61.

Cheng, H. T. (1988). Algorithms for partially observable Markov decision processes. PhD thesis, University of British Columbia, Vancouver, BC, Canada.

Littman, M. L., Cassandra, A. R., & Kaelbling, L. P. (1995). Efficient dynamic-programming updates in partially observable Markov decision processes. TR CS-95-19, Department of Computer Science, Brown University, Providence, Rhode Island 02912, USA.

G. E. Monahan (1982), A survey of partially observable Markov decision processes: theory, models, and algorithms, *Management Science*, 28 (1), pp. 1-16.

Sondik, E. J. (1971). The optimal control of partially observable Markov processes. PhD thesis, Stanford University, Stanford, California, USA.

White III, C. C. (1991). Partially observed Markov decision processes: A survey. *Annals of Operations Research*, 32.

N. L. Zhang and W. Liu (1997), A model approximation scheme for planning in stochastic domains, *Journal of Artificial Intelligence Research*, 7, pp. 199-230.